\documentclass{article} 
\usepackage{iclr2022_conference,times}

\usepackage{amsmath,amsfonts,bm}

\def\eqref#1{equation~\ref{#1}}

\def\1{\bm{1}}

\DeclareMathAlphabet{\mathsfit}{\encodingdefault}{\sfdefault}{m}{sl}
\SetMathAlphabet{\mathsfit}{bold}{\encodingdefault}{\sfdefault}{bx}{n}

\usepackage{hyperref}
\usepackage{url}

\usepackage{times}
\usepackage{latexsym}

\usepackage[T1]{fontenc}

\usepackage[utf8]{inputenc}

\usepackage{microtype}

\usepackage{amsfonts}       
\usepackage{amsmath}

\usepackage{appendix}
\usepackage{hhline}
\usepackage{multirow}
\usepackage{blindtext}
\usepackage{makecell}

\usepackage{soul}
\usepackage{booktabs}
\usepackage{nicefrac}       
\usepackage{lipsum} 
\usepackage{stmaryrd}

\usepackage{amssymb}
\usepackage{subfigure}
\usepackage{mathtools, cuted}
\usepackage{rotating}
\usepackage{adjustbox}  

\usepackage{wrapfig}

\title{KELM: Knowledge Enhanced Pre-Trained Language Representations with Message Passing on Hierarchical Relational Graphs}

\author{Yinquan Lu\textsuperscript{\rm 1,}\textsuperscript{\rm 5}\thanks{This work is done when Yinquan Lu, Haonan Lu and
Guirong Fu work at Huawei Technologies Co., Ltd.}\quad
Haonan Lu\textsuperscript{\rm 2}\footnotemark[1]\ \ \thanks{Corresponding author}\quad
Guirong Fu\textsuperscript{\rm 3}\footnotemark[1]\quad
Qun Liu\textsuperscript{\rm 4}\\
Huawei Technologies Co., Ltd.\textsuperscript{\rm 1} \quad
OPPO Guangdong Mobile Telecommunications Co., Ltd.\textsuperscript{\rm 2} \\
ByteDance\textsuperscript{\rm 3} \quad
Huawei Noah’s Ark Lab\textsuperscript{\rm 4} \quad
Shanghai AI Laboratory\textsuperscript{\rm 5} \\
\texttt{luyinquan@pjlab.org.cn,}\quad
\texttt{luhaonan@oppo.com}\\
\texttt{fuguirong@bytedance.com,}\quad
\texttt{qun.liu@huawei.com}
}

\iclrfinalcopy 
\begin{document}

\maketitle

\begin{abstract}
Incorporating factual knowledge into pre-trained language models (PLM) such as BERT is an emerging trend in recent NLP studies. However, most of the existing methods combine the external knowledge integration module with a modified pre-training loss and re-implement the pre-training process on the large-scale corpus. Re-pretraining these models is usually resource-consuming, and difficult to adapt to another domain with a different knowledge graph (KG). Besides, those works either cannot embed knowledge context dynamically according to textual context or struggle with the knowledge ambiguity issue. In this paper, we propose a novel knowledge-aware language model framework based on fine-tuning process, which equips PLM with a unified knowledge-enhanced text graph that contains both text and multi-relational sub-graphs extracted from KG. We design a hierarchical relational-graph-based message passing mechanism, which allows the representations of injected KG and text to mutually update each other and can dynamically select ambiguous mentioned entities that share the same text$\footnote{Words or phrases in the text corresponding to certain entities in KGs are often named \textbf{``entity mentions''}. While entities in KGs that correspond to entity mentions in the text are often named \textbf{``mentioned entities''}}$. Our empirical results show that our model can efficiently incorporate world knowledge from KGs into existing language models such as BERT, and achieve significant improvement on the machine reading comprehension (MRC) tasks compared with other knowledge-enhanced models.
\end{abstract}

\section{Introduction}
Pre-trained language models benefit from the large-scale corpus and can learn complex linguistic representation~\citep{devlin-etal-2019-bert, liu2019roberta, yang2020xlnet}. Although they have achieved promising results in many NLP tasks, they neglect to incorporate structured knowledge for language understanding.
Limited by implicit knowledge representation, existing PLMs are still difficult to learn world knowledge efficiently~\citep{Poerner2019BERTIN, yu2020jaket}. For example, hundreds of related training samples in the corpus are required to understand the fact ``\(ban\) means an official prohibition or edict against something'' for PLMs. 

By contrast, knowledge graphs (KGs) explicitly organize the above fact as a triplet \textit{``(ban, hypernyms, prohibition)''}. Although domain knowledge can be represented more efficiently in KG form, entities with different meanings share the same text may happen in a KG (knowledge ambiguity issue). For example, one can also find \textit{``(ban, hypernyms, moldovan monetary unit)''} in WordNet~\citep{Miller95wordnet:a}. 
Recently, many efforts have been made on leveraging heterogeneous factual knowledge in KGs to enhance PLM representations. These models generally adopt two methods: (1). Injecting pre-trained entity embeddings into PLM explicitly, such as ERNIE~\citep{zhang2019ernie}, which injects entity embeddings pre-trained on a knowledge graph by using TransE~\citep{Bordes2013TranslatingEF}. 
(2). Implicitly learning factual knowledge by adding extra pre-training tasks such as entity-level mask, entity-based replacement prediction, etc.~\citep{wang2020kadapter, sun2020colake}. Some studies use both of the above two methods such as CokeBERT~\citep{su2020cokebert}. 

However, as summarized in Table~\ref{model comparison v1} of Appendix, most of the existing knowledge-enhanced PLMs need to re-pretrain the models based on an additional large-scale corpus, they mainly encounter two problems below: 
(1) Incorporating external knowledge during pretraining is usually resource-consuming and difficult to adapt to other domains with different KGs. By checking the third column of Table~\ref{model comparison v1} in Appendix, one can see that most of the pretrain-based models use Wiki-related KG as their injected knowledge source. These models also use English Wikipedia as pre-training corpus. They either use an additional entity linking tool (e.g. TAGME~\citep{Ferragina2010TAGMEOA}) to align the entity mention in the text to a single mentioned entity in a Wiki-related KG uniquely or directly treat hyperlinks in Wikipedia as entity annotations. These models depend heavily on the one-to-one mapping relationship between Wikipedia corpus and Wiki-related KG, thus they never consider handling knowledge ambiguity issue.
(2) These models with explicit knowledge injection usually use algorithms like BILINEAR~\citep{yang2015embedding} to obtain pre-trained KG embeddings, which contain information about graph structure. Unfortunately, their knowledge context is usually static and cannot be embedded dynamically according to textual context.

Several works~\citep{qiu-etal-2019-machine, yang-etal-2019-enhancing-pre} concentrate on injecting external knowledge based on fine-tuning PLM on downstream tasks, which is much easier to change the injected KGs and adapt to relevant domain tasks. They either cannot consider multi-hop relational information, or struggle with knowledge ambiguity issue. How to fuse heterogeneous information dynamically based on the fine-tuning process on the downstream tasks and use the information of injected KGs more efficiently remains a challenge.


\begin{wrapfigure}{r}{8.3cm}
\centering  
\includegraphics[height=38mm]{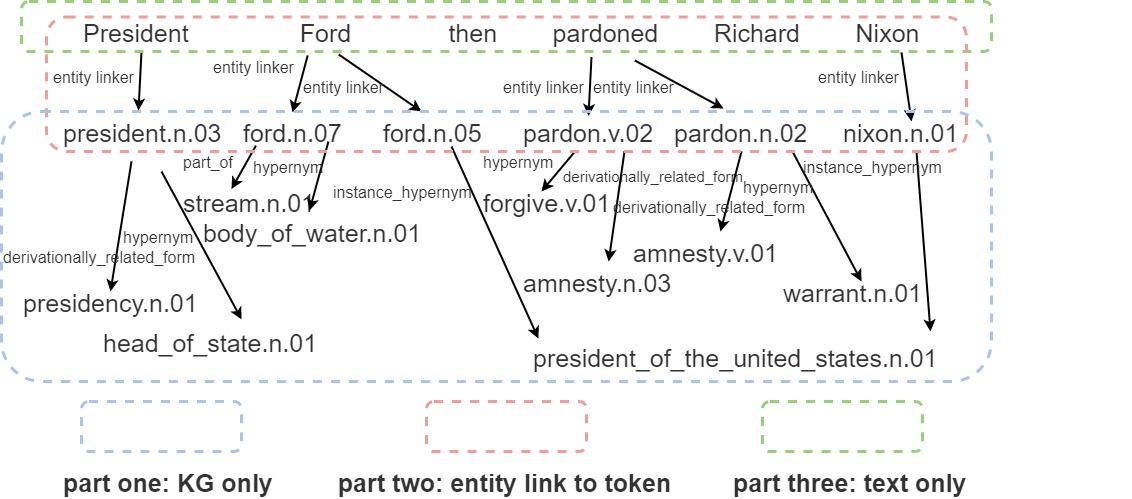} 

\caption{Unified Knowledge-enhanced Text Graph (UKET) consists of three parts corresponding to our model: (1) KG only part, (2) Entity link to token graph, (3) Text only graph.}
\label{KWG}
\end{wrapfigure}

To overcome the challenges mentioned above, we propose a novel framework named \textbf{KELM}, which injects world knowledge from KGs during the fine-tuning phase by building a \textbf{Unified Knowledge-enhanced Text Graph (UKET)} that contains both injected sub-graphs from external knowledge and text. The method extends the input sentence by extracting sub-graphs centered on every mentioned entity from KGs. In this way, we can get a Unified Knowledge-enhanced Text Graph as shown in Fig.~\ref{KWG}, which is made of three kinds of graph: (1) The injected knowledge graphs, referred to as the ``\textbf{KG only}'' part; (2) The graph about entity mentions in the text and mentioned entities in KGs, referred to as the ``\textbf{entity link to token}'' part. Entity mentions in the text are linked with mentioned entities in KGs by string matching, so one entity mention may trigger several mentioned entities that share the same text in the injected KGs (e.g. “Ford” in Fig.~\ref{KWG}); (3) The ``\textbf{text only}'' part, where the input text sequence is treated as a fully-connected word graph just like classical Transformer architecture~\citep{vaswani2017attention}.

\begin{figure*}[!htb]
\centering  
\includegraphics[width=0.7\linewidth]{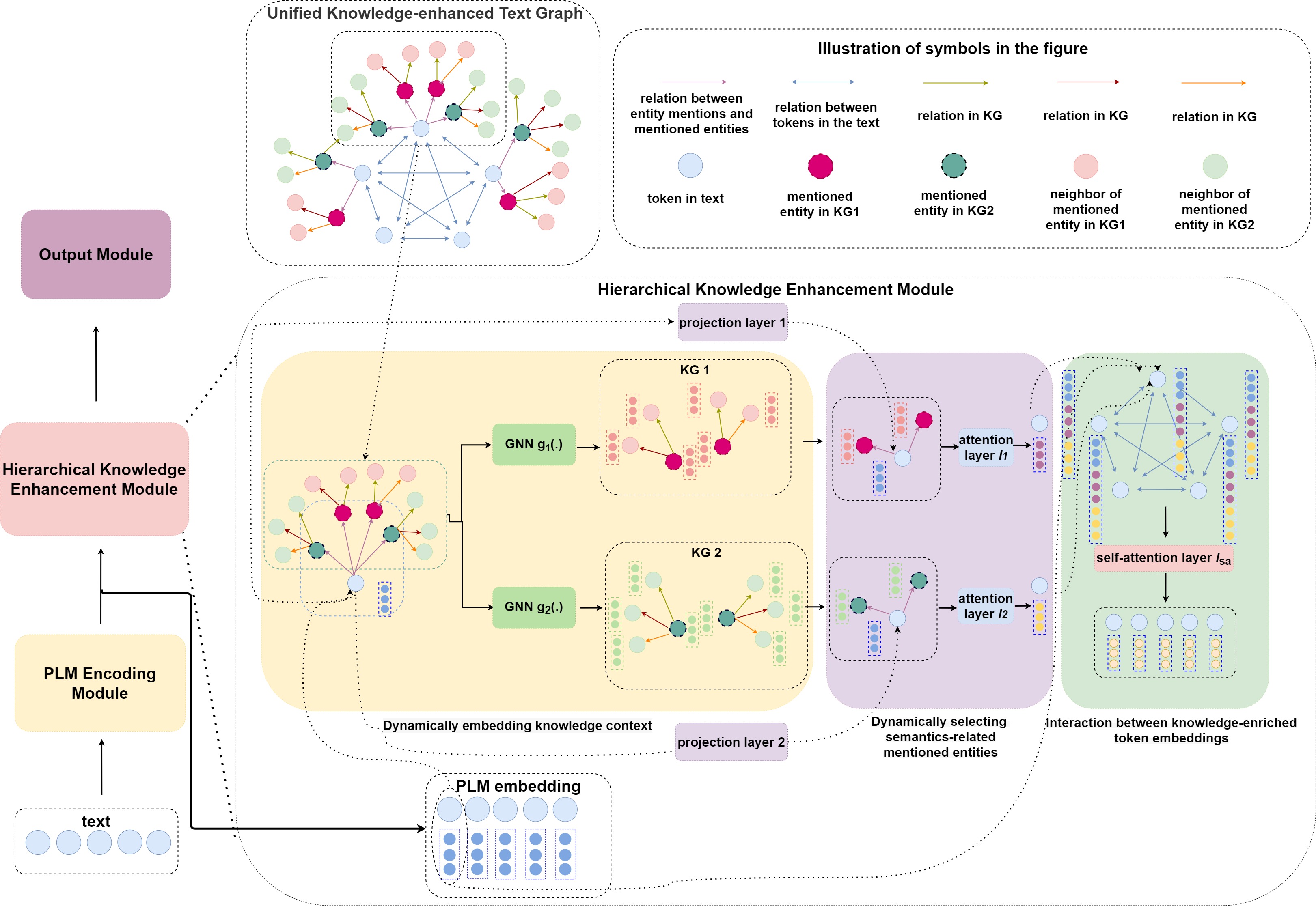}
\caption{Framework of KELM (left) and illustrates how to generate knowledge-enriched token embeddings (right).}
\label{module}
\end{figure*}

Based on this unified graph, we design a novel Hierarchical relational-graph-based Message Passing (HMP) mechanism to fuse heterogeneous information on the output layer of PLM. The implementation of HMP is via a Hierarchical Knowledge Enhancement Module as depicted in Fig.~\ref{module}, which also consists of three parts, and each part is designed for solving the different problems above: 
(1) For reserving the structure information and dynamically embedding injected knowledge, we utilize a relational GNN (e.g. rGCN~\citep{schlichtkrull2017modeling}) to aggregate and update representations of extracted sub-graphs for each injected KG (corresponding to the ``KG only'' part of UKET). All mentioned entities and their K-hop neighbors in sub-graphs are initialized by pre-trained vectors obtained from the classical knowledge graph embedding (KGE) method (we adopt BILINEAR here). In this way, knowledge context can be dynamically embedded, the structural information about the graph is also kept;
(2) For handling knowledge ambiguity issue and selecting relevant mentioned entities according to the input context, we leverage a specially designed attention mechanism to weight these ambiguous mentioned entities by using the textual representations of words/tokens to query the representations of their related mentioned entities in KGs (corresponding to the ``entity link to token'' graph of UKET). The attention score can help to select knowledge according to the input sentence dynamically. By concatenating the outputs of this step with the original outputs of PLM, we can get a knowledge-enriched representation for each token;
(3) For further interactions between knowledge-enriched tokens, we employ a self-attention mechanism that operates on the fully-connected word graph (corresponding to the ``text only'' graph of UKET) to allow the knowledge-enriched representation of each token to further interact with others.

We conduct experiments on the MRC task, which requires a system to comprehend a given text and answer questions about it. 
In this paper, to prove the generalization ability of our method, we evaluate KELM on both the extractive-style MRC task (answers can be found in a span of the given text) and the multiple-response-items-style MRC task (each question is associated with several choices for answer-options, the number of correct answer-options is not pre-specified).
MRC is a challenging task and represents a valuable path towards natural language understanding (NLU). With the rapid increment of knowledge, NLU becomes more difficult since the system needs to absorb new knowledge continuously. Pre-training models on large-scale corpus is inefficient. Therefore, fine-tuning the knowledge-enhanced PLM on the downstream tasks directly is crucial in the application.$\footnote{Code is available at ~\href{https://github.com/nlp-anonymous-happy/anonymous-KG-guided-NLP}{here}.}$

\section{Related Work}
\subsection{Knowledge Graph Embedding}
We denote a directed knowledge graph as \(G(\mathcal{E},\mathcal{R})\), where \(\mathcal{E}\) and \(\mathcal{R}\) are sets of entities and relations, respectively. We also define \(\mathcal{F}\) as a set of facts, a fact stored in a KG can be expressed as a triplet \((h, r, t) \in \mathcal{F}\), which indicates a relation \(r\) pointing from the head entity \(h\) to tail entity \(t\), where \(h, t \in \mathcal{E}\) and \(r \in \mathcal{R}\). KGE aims to extract topological information in KG and to learn a set of low-dimensional representations of entities and relations by knowledge graph completion task~\citep{yang2015embedding, lu2020dense}. 

\subsection{Multi-relational Graph Neural Network}
Real-world KGs usually include several relations. However, traditional GNN models such as GCN~\citep{kipf2017semisupervised}, and GAT~\citep{GAT} can only be used in the graph with one type of relation. ~\citep{schlichtkrull2017modeling, haonan2019graph} generalizes traditional GNN models by performing relation-specific aggregation, making it possible to encode relational graphs. The use of multi-relational GNN makes it possible to encode injected knowledge embeddings dynamically in SKG and CokeBERT.

\subsection{Joint Language and Knowledge Models}
Since BERT was published in 2018, many efforts have been made for further optimization, basically focusing on the design of the pre-training process and the variation of the encoder. For studies of knowledge-enhanced PLMs, they also fall into the above two categories or combine both of them sometimes. Despite their success in leveraging external factual knowledge, the gains are limited by computing resources, knowledge ambiguity issue, and the expressivity of their methods for the fusion of heterogeneous information, as summarized in Table~\ref{model comparison v1} of Appendix and the introduction part.

Recent studies notice that the architecture of Transformer treats input sequences as fully-connected word graphs, thus some of them try to integrate injected KGs and textual context into a unified data structure. Here we argue that UKET in our KELM is different from the WK graph proposed in CoLAKE/K-BERT. These two studies heuristically convert textual context and entity-related sub-graph into input sequences, both entities and relations are treated as input words of the PLM, then they leverage a Transformer with a masked attention mechanism to encode those sequences from the embedding layer and pre-train the model based on the large-scale corpus. Unfortunately, it is not trivial for them to convert the second or higher order neighbors related to textual context~\citep{su2020cokebert}, the structural information about the graph is lost. UKET differs from the WK graph of CoLAKE/K-BERT in that, instead of converting mentioned entities, relations, and text into a sequence of words and feeding them together into the input layer of PLM (they unify text and KG into a sequence), UKET unifies text and KG into a graph. Besides, by using our UKET framework, the knowledge fusion process of KELM is based on the representation of the last hidden layer of PLM, making it possible to directly fine-tune the PLM on the downstream tasks without re-pretraining the model. SKG also utilizes relational GNN to fuse information of KGs and text representation encoded by PLM. However, SKG only uses GNN to dynamically encode the injected KGs, which corresponds to part one of Fig.~\ref{KWG}. Outputs of SKG are made by directly concatenating outputs of graph encoder with the outputs of PLM. It cannot select ambiguous knowledge and forbids the interactions between knowledge-enriched tokens corresponding to part two and part three of Fig.~\ref{KWG}, respectively. KT-NET uses a specially designed attention mechanism to select relevant knowledge from KGs. For example, it treats all synsets of entity mentions within the WN18$\footnote{A subset of WordNet.}$ as candidate KB concepts.
This limits the ability of KT-NET to select the most relevant mentioned entities$\footnote{Refer the example given in the case study of KT-NET, the most relevant concept for the word ``ban'' is ``forbidding\_NN\_1'' (with the probability of 86.1\%), but not ``ban\_NN\_4''.}$. Moreover, the representations of injected knowledge are static in KT-NET, they cannot dynamically change according to textual context, the information about the original graph structure in KG is also lost.

\section{Methodology}

The architecture of KELM is shown in Fig.~\ref{module}. It consists of three main modules: (1) PLM Encoding Module; (2) Hierarchical Knowledge Enhancement Module; (3) Output Module. 

\subsection{PLM Encoding Module}

This module utilizes PLM (e.g.BERT) to encode text to get textual representations for passages and questions. An input example of the MRC task includes a paragraph and a question with a candidate answer, represented as a single sequence of tokens of the length \(n\): \(T\)=\(\left\{[CLS], Q, (A), [SEP], P, [SEP]\right\}\)=\(\left\{t_i\right\}_{i=1}^{n}\), where \(Q\), \(A\) and \(P\) represent all tokens for question, candidate answer and paragraph, respectively$\footnote{Depending on the type of MRC task (extractive-style v.s. multiple-response-items-style), candidate answer A is not required in the sequence of tokens for the extractive-style MRC task.}$. \([SEP]\) and \([CLS]\) are special tokens in BERT and defined as a sentence separator and a classification token, respectively. \(i\)-th token in the sequence is represented by \(\vec{h}_{i}^{t} \in \mathbb{R}^{d_t}\), where \(d_t\) is the last hidden layer size of used PLM.

\subsection{Hierarchical Knowledge Enhancement Module}
This module is the implementation of our proposed HMP mechanism to fuse information of textual and graph context. We will formally introduce graph construction for UKET, and the three sub-processes of HMP in detail in the following sections. 

\subsubsection{Construction of UKET Graph}
(1) Given a set with \(|\mathcal{Q}|\) elements: \(\left\{G_{k}^{q}(\mathcal{E}_{k}^{q},\mathcal{R}_{k}^{q})\right\}_{q=1}^{|\mathcal{Q}|}\) and input text, where \(|\mathcal{Q}|\) is the total number of injected KGs, and \(q\) indicates the \(q\)-th KG. We denote the set of entity mentions related to the \(q\)-th KG as \(\mathcal{X}^{q}\)=\(\{x_{i}^{q}\}_{i=1}^{|\mathcal{X}^{q}|}\), where \(|\mathcal{X}^{q}|\) is the number of entity mentions in the text. 
The corresponding mentioned entities are shared by all tokens in the same entity mention. All mentioned entities \(\mathcal{M}^{q}\)=\(\{m_{i}^{q}\}_{i=1}^{|\mathcal{M}^{q}|}\) are linked with their relevant entity mentions in the text, where \(|\mathcal{M}^{q}|\) is the number of mentioned entities in the \(q\)-th KG. We define this "entity link to token graph" in Fig.~\ref{KWG} as \( G_{m}^{q}(\mathcal{E}_{m}^{q}, \mathcal{R}_{m}^{q})\), where \(\mathcal{E}_{m}^{q}\)=\(\mathcal{X}^{q}\cup\mathcal{M}^{q}\) is the union of entity mentions and their relevant mentioned entities, \(\mathcal{R}_{m}^{q}\) is a set with only one element that links mentioned entities and their relevant entity mentions. 
(2) For \(i\)-th mentioned entity \(m_i^{q}\) in \( \mathcal{M}^{q}\), we retrieve all its K-hop neighbors \(\{\mathcal{N}_{m_{i}^{q}}^{x}\}_{x=0}^{K}\) from the \(q\)-th knowledge graph, where \(\mathcal{N}_{m_{i}^{q}}^{x}\) is a set of \(i\)-th mentioned entity's x-hop neighbors, hence we have \(\mathcal{N}_{m_{i}^{q}}^{0}\)=\(\{m_i^{q}\}\). We define "KG-only graph": \(G_s^{q}(\mathcal{E}_{s}^{q}, \mathcal{R}_{s}^{q})\), where \(\mathcal{E}_{s}^{q}\)=\(\bigcup_{i=0}^{|\mathcal{M}^{q}|}\bigcup_{x=0}^{K}\mathcal{N}_{m_{i}^{q}}^{x}\) is the union of all mentioned entities and their neighbors within the K-hops sub-graph, and \(\mathcal{R}_{s}^{q}\) is a set of all relations in the extracted sub-graph of \(q\)-th KG. 
(3) The text sequence can be considered as a fully-connected word graph as pointed out previously. This ``text-only graph'' can be denoted as \(G_t(\mathcal{E}_{t}, \mathcal{R}_{t})\), where \(\mathcal{E}_{t}\) is all tokens in text and \(\mathcal{R}_{t}\) is a set with only one element that connects all tokens. 
Finally, we define the full hierarchical graph consisting of all three parts \(\left\{G_{s}^{q}\right\}_{q=1}^{|\mathcal{Q}|}\), \(\left\{G_{m}^{q}\right\}_{q=1}^{|\mathcal{Q}|}\), and \(G_t\), as Unified Knowledge-enhanced Text Graph (UKET).

\subsubsection{Dynamically Embedding Knowledge Context}
We use pre-trained vectors obtained from the KGE method to initialize representations of entities in \(G_s^{q}(\mathcal{E}_{s}^{q}, \mathcal{R}_{s}^{q})\). Considering the structural information of injected knowledge graph forgotten during training, we utilize \(|\mathcal{Q}|\) independent GNN encoders (i.e. \(g_{1}(.)\), \(g_{2}(.)\) in Fig.~\ref{module}, which is the case of injecting two independent KGs in our experiment setting) to dynamically update entity embeddings of \(|\mathcal{Q}|\) injected KGs. We use rGCN to model the multi-relational nature of the knowledge graph. To update \(i\)-th node of \(q\)-th KG in \(l\)-th rGCN layer:
\begin{equation}
\begin{aligned}
\vec{s}_{i}^{\ q(l+1)}=\sigma (\sum\limits_{r\in \mathcal{R}_{s}^{q}} \sum\limits_{j\in \mathcal{N}_{i}^{r}} \frac{1}{|\mathcal{N}_{i}^{r}|} W_{r}^{q(l)} \vec{s}_{j}^{\ q(l)})
\end{aligned}
\end{equation}

Where \(\mathcal{N}_{i}^{r}\) is a set of neighbors of \(i\)-th node under relation \(r \in \mathcal{R}_{s}^{q}\). \(W_{r}^{q(l)}\) is trainable weight matrix at \(l\)-th layer and \(\vec{s}_{i}^{\ q(l+1)}\) is the hidden state of \(i\)-th node at (\(l+\)1)-th layer.
After \(L\) updates, \(|\mathcal{Q}|\) sets of node embeddings are obtained. The output of the \(q\)-th KG can be represented as \(S^{q} \in \mathbb{R}^{|\mathcal{E}_{s}^{q}| \times d_{q}}\), where \(|\mathcal{E}_{s}^{q}|\) and \(d_{q}\) are the numbers of nodes of extracted sub-graph and the dimension of pre-trained KGE, respectively.

\subsubsection{Dynamically Selecting Semantics-Related Mentioned Entities}
To handle the knowledge ambiguity issue, we introduce an attention layer to weight these ambiguous mentioned entities by using the textual representations of tokens (outputs of Section 3.1) to query their semantics-related mentioned entities representations in KGs. Here, we follow the attention mechanism of GAT to update each entity mention embedding in \(G_{m}^{q}\): 
\begin{equation}
\begin{aligned}
\vec{x}_{i}^{\ q}=\sigma (\sum\limits_{j\in \mathcal{N}_{i}^{q}} \alpha_{ij}^{q}W_{q}\vec{s}_{j}^{\ q})
\end{aligned}
\label{hierarchy}
\end{equation}
\noindent Where \(\vec{s}_{j}^{\ q}\) is the output embeddings from the \(q\)-th rGCN in the previous step. \(\vec{x}_{i}^{q} \) is the hidden state of \(i\)-th entity mention \(x_{i}^{q}\) in \(\mathcal{X}^{q}\), and \(\mathcal{N}_{i}^{q}\) is a set of neighbors of \(x_{i}^{q}\) in \(G_{m}^{q}\). \(W_q \in \mathbb{R}^{d_{out} \times d_{in}}\) is a trainable weight matrix, we set \(d_{in}\)=\(d_{out}\)=\(d_{q}\) (thus \(\vec{x}_{i}^{\ q} \in \mathbb{R}^{d_{q}}\)). \(\sigma\) is a nonlinear activation function. \(\alpha_{ij}^{q}\) is the attention score that weights ambiguous mentioned entities in the \(q\)-th KG:

\begin{equation}
\begin{aligned}
\alpha_{ij}^{q}=\frac{exp(\text{LeakyReLU}(\vec{\alpha}^{T}_{q}[W_{q}\vec{h}_{i}^{t'} || W_{q}\vec{s}_{j}^{\ q}]))}{\sum\limits_{k\in \mathcal{N}_{i}^{q}}exp(\text{LeakyReLU}(\vec{\alpha}^{T}_{q}[W_{q}\vec{h}_{i}^{t'} || W_{q}\vec{s}_{k}^{\ q}]))}
\label{(6)}
\end{aligned}
\end{equation}

\noindent The representation \(\vec{h}_{i}^{t}\) with a dimension of \(d_t\) is projected to the dimension of \(d_{q}\), before using it to query the related mentioned entity embeddings of \( S^{q}\): \(\vec{h}_{i}^{t'} = W_{proj}^{q} \vec{h}_{i}^{t}\), where \(W_{proj}^{q} \in \mathbb{R}^{d_{q} \times d_t}\). \( \vec{\alpha}_{q} \in \mathbb{R}^{2d_{q}}\) is a trainable weight vector. \(\cdot^{T}\) is the transposition operation and \(||\) is the concatenation operation.

Finally, we concatenate outputs of \(|\mathcal{Q}|\) KGs with textual context representation to get final knowledge-enriched representation:
\begin{equation}
\begin{aligned}
\vec{h}_{i}^{k}=[\vec{h}_{i}^{t}, \vec{x}_{i}^{1}, \dots, \vec{x}_{i}^{|\mathcal{Q}|}] \in \mathbb{R}^{d_t + d_{1} + \dots + d_{|\mathcal{Q}|}}
\label{(7)}
\end{aligned}
\end{equation}
If token \(t_i\) can't match any entity in \(q\)-th KG (say \(t_i \notin \mathcal{X}^{q}\)), we fill \(\vec{x}_{i}^{\ q}\) in Eq.\ref{(7)} with zeros. Note that mentioned entities in KGs are not always useful, to prevent noise, we follow ~\citep{yang-mitchell-2017-leveraging}'s work and add an extra sentinel node linked to each entity mention in \(G_{m}^{q}\). The sentinel node is initialized by zeros and not trainable, which is the same as the case of no retrieved entities in the KG. In this way, according to the textual context, KELM can dynamically select mentioned entities and avoid introducing knowledge noise.

\subsubsection{Interaction Between Knowledge-enriched Token Embeddings}
To allow knowledge-enriched tokens' representations to propagate to each other in the text, we use a fully-connected word graph \(G_t\), with knowledge-enriched representations from outputs of the previous step, and employ the self-attention mechanism similar to KT-NET to update token's embedding. The final representation for \(i\)-th token in the text is \(\vec{h}_i^f \in \mathbb{R}^{6 \ast (d_t+d_{1} + \dots + d_{|\mathcal{Q}|})}\).

\subsection{Output Module}
\subsubsection{Extractive-style MRC task} 
A simple linear transformation layer and softmax operation are used to predict start and end positions of answers. For \(i\)-th token, the probabilities to be the start and end position of answer span are:
\(p_{i}^{s} = \frac{exp(w_{s}^{T} \vec{h}_i^f)}{\sum\limits_{j=1}^{n} exp(w_{s}^{T} \vec{h}_j^f)}, p_{i}^{e} = \frac{exp(w_{e}^{T} \vec{h}_i^f)}{\sum\limits_{j=1}^{n} exp(w_{e}^{T} \vec{h}_j^f)}\), where \(w_{s}, w_{e} \in \mathbb{R}^{6 \ast (d_t+d_{1} + \dots + d_{|\mathcal{Q}|})}\) are trainable vectors and \(n\) is the number of tokens. The training loss is calculated by the log-likelihood of the true start and end positions: \(\mathcal{L}=-\frac{1}{N}\sum\limits_{i=1}^{N}(log\, p_{y_{i}^{s}}^{s} + log\, p_{y_{i}^{e}}^{e})\),
\noindent where \(N\) is the total number of examples in the dataset, \( y_i^s\) and \( y_i^e\) are the true start and end positions of \(i\)-th query's answer, respectively. During inference, we pick the span \((a, b)\) with maximum \(p_{a}^{s}p_{b}^{e}\) where \(a \leq b\) as predicted anwser. 

\subsubsection{Multiple-response-items-style MRC task}
Since answers to a given question are independent of each other, to predict the correct probability of each answer, a fully connected layer followed by a sigmoid function is applied on the final representation of \([CLS]\) token in BERT. 

\section{Experiments}
\subsection{Datasets}
In this paper, we empirically evaluate KELM on both two types of MRC benchmarks in SuperGLUE~\citep{wang2020superglue}: \textbf{ReCoRD}~\citep{zhang2018record} (extractive-style) and \textbf{MultiRC}~\citep{MultiRC2018} (multiple-response-items-style). Detailed descriptions of the two datasets can be found in Appendix B. On both datasets, the test set is not public, one has to submit the predicted results to the organization to get the final test score. Since frequent submissions to probe the unseen test set are not encouraged, we only submit our best model once for each of the datasets, thus the statistics of the results (e.g., mean, variance, etc.) are not applicable. We use Exact Match (EM) and (macro-averaged) F1 as the evaluation metrics.


\noindent\textbf{External Knowledge} We adopt knowledge sources the same as used in KT-NET: WordNet and NELL~\citep{Carlson10}. Representations of injected knowledge are initialized by resources provided by~\citep{yang-mitchell-2017-leveraging}. The size of these embeddings is 100. We retrieve related knowledge from the two KGs in a given sentence and construct UKET graph (as shown in Section 3.2.1). More details about entity embedding and concepts retrieval are available in Appendix B.

\subsection{Experimental Setups}

\textbf{Baselines and Comparison Setting} 
Because we use \(\text{BERT}_{\text{large}}\) as the base model in our method, we use it as our primary baseline for all tasks. 
For fair comparison, we mainly compare our results with two fine-tune-based knowledge-enhanced models: KT-NET and SKG, which also evaluate their results on ReCoRD with \textbf{\(\text{BERT}_{\text{large}}\)} as the encoder part. 
As mentioned in the original paper of KT-NET, KT-NET mainly focuses on the extractive-style MRC task. We also evaluate KT-NET on the multiple-response-items-style MRC task and compare the results with KELM.
We evaluate our approach in three different KB settings: \(\text{KELM}_{\text{WordNet}}\), \(\text{KELM}_{\text{NELL}}\), and \(\text{KELM}_{\text{Both}}\), to inject KG from WordNet, NELL, and both of the two, respectively (The same as KT-NET). 
Implementation details of our model are presented in Appendix C.

\begin{table}[!ht]
\begin{minipage}{0.48\linewidth}

\centering
\scriptsize
\setlength{\tabcolsep}{1.0mm}
 \begin{tabular}{clcccc}
  \toprule
    \multicolumn{2}{c}{} & \multicolumn{2}{c}{\textbf{Dev}} & \multicolumn{2}{c}{\textbf{Test}}  \\
  	&	\textbf{Model} & \textbf{EM}	&	\textbf{F1} &	\textbf{EM}	&	\textbf{F1} \\
  \midrule
    ~	&	\(\text{BERT}_{\text{large}}\)	&	70.2	&	72.2	&	71.3	&	72.0	\\
    \midrule															
    ~	&	SKG+\(\text{BERT}_{\text{large}}\)	&	70.9	&	71.6	&	72.2	&	72.8	\\
    ~	&	\(\text{KT-NET}_{\text{WordNet}}\)	&	70.6	&	72.8	&	-	&	-	\\
    ~	&	\(\text{KT-NET}_{\text{NELL}}\)	&	70.5	&	72.5	&	-	&	-	\\
    ~	&	\(\text{KT-NET}_{\text{BOTH}}\)	&	71.6	&	73.6	&	73.0	&	74.8	\\
    \midrule	
    
    ~	&	\(\text{KELM}_{\text{WordNet}}\)	&	\textbf{75.4}	&	\textbf{75.9}	&	\underline{75.9}	&	\underline{76.5}	\\
    ~	&	\(\text{KELM}_{\text{NELL}}\)	&	74.8	&	75.3	&	\underline{75.9}	&	76.3	\\
    ~	&	\(\text{KELM}_{\text{Both}}\)	&	\underline{75.1}	&	\underline{75.6}	&	\textbf{76.2}	&	\textbf{76.7}	\\
\bottomrule
\end{tabular}
\caption{Result on ReCoRD.}
\label{main results table extractive}

\end{minipage}\begin{minipage}{0.48\linewidth}

\setlength{\belowcaptionskip}{-5mm}
\centering
\scriptsize
\setlength{\tabcolsep}{1.0mm}
 \begin{tabular}{clcccc}
  \toprule
    \multicolumn{2}{c}{} & \multicolumn{2}{c}{\textbf{Dev}} & \multicolumn{2}{c}{\textbf{Test}}  \\
  	&	\textbf{Model} & \textbf{EM}	&	\textbf{F1} &	\textbf{EM}	&	\textbf{F1} \\
  \midrule
    ~	&	\(\text{BERT}_{\text{large}}\)	&	-	&	-	&	24.1	&	70.0	\\
    \midrule															
    ~	&	\(\text{KT-NET}_{\text{BOTH}}^{*}\)	&	26.7	&	\textbf{71.7}	&	25.4 & \textbf{71.1}	\\
    \midrule	
    
    ~	&	\(\text{KELM}_{\text{WordNet}}\)	&	\underline{29.2}	&	70.6	&	25.9	&	69.2	\\
    ~	&	\(\text{KELM}_{\text{NELL}}\)	&	27.3	&	70.4	&	\underline{26.5}	&	70.6	\\
    ~	&	\(\text{KELM}_{\text{Both}}\)	&	\textbf{30.3}	&	\underline{71.0}	&	\textbf{27.2}	&	\underline{70.8}	\\

\bottomrule
\end{tabular}
\caption{Result on MultiRC. [*] are from our implementation.}
\label{main results table multiple}
\end{minipage}

\end{table}

\subsection{Results}
The results for the extractive-style MRC task and multiple-response-items-style MRC task are given in Table~\ref{main results table extractive} and Table~\ref{main results table multiple}, respectively. The scores of other models are taken directly from the leaderboard of SuperGLUE$\footnote{\url{https://super.gluebenchmark.com/leaderboard} (Nov.14th, 2021)}$ and literature~\citep{qiu-etal-2019-machine, yang-etal-2019-enhancing-pre}. 
In this paper, our implementation is based on a single model, and hence comparing with ensemble based models is not considered. 
Best results are labeled in bold and the second best are underlined.

Results on the \textbf{dev set} of ReCoRD show that: (1) KELM outperforms \(\text{BERT}_{\text{large}}\), irrespective of which external KG is used. Our best KELM offers a \textbf{5.2/3.7} improvement in EM/F1 over \(\text{BERT}_{\text{large}}\). (2) KELM outperforms previous SOTA knowledge-enhanced PLM (KT-NET) by \textbf{+3.8 EM/+2.3 F1}. In addition, KELM outperforms KT-NET significantly in all three KB settings. On the \textbf{dev set} of MultiRC, the best KELM offers a \textbf{3.6} improvement in EM over KT-NET. Although the performance on F1 drop a little compared with KT-NET, we still get a gain of \textbf{+2.9 (EM+F1)} over the former SOTA model$\footnote{The best model is chosen according to the EM+F1 score (same as KT-NET).}$. 

Results on the \textbf{test set} further demonstrate the effectiveness of KELM and its superiority over the previous works. 
On ReCoRD, it significantly outperforms the former SOTA knowledge-enhanced PLM (finetuning based model) by \textbf{+3.2 EM/+1.9 F1}. And on MultiRC, KELM offers a \textbf{3.1/0.8} improvement in EM/F1 over \(\text{BERT}_{\text{large}}\), and achieves a gain of \textbf{+1.5 (EM+F1)} over KT-NET.

\section{Case Study}
This section uses an example in ReCoRD to show how KELM avoids knowledge ambiguity issue and selects the most relevant mentioned entities adaptively w.r.t the textual context. Recall that given a token \(t_i\), the importance of a mentioned entity \(m_{j}^{q}\) in \(q\)-th KG is scored by the attention weight \(\alpha_{ij}^{q}\) in Eq.\ref{hierarchy}. To illustrate how KELM can select the most relevant mentioned entities, we analyze the example that was also used in the case study part of KT-NET. The question of this example is ``\textit{Sudan remains a XXX-designated state sponsor of terror and is one of six countries subject to the Trump administration’s \textbf{ban}}'', where the ``XXX'' is the answer that needs to be predicted. The case study in KT-NET shows the top 3 most relevant concepts from WordNet for the word ``ban'' are ``forbidding.n.01'', ``proscription.n.01'', and ``ban.v.02'', with the weights of 0.861, 0.135, and 0.002, respectively. KT-NET treats all synsets of a word as candidate KG concepts, both ``forbidding.n.01'' and ``ban.v.02'' will be the related concepts of the word ``ban'' in the text. Although KT-NET can select relevant concepts and suppress the knowledge noise through its specially designed attention mechanism, we still observe two problems from the previous case study: (1) KT-NET cannot select the most relevant mentioned entities in KG that share the same string in the input text. (2) Lack of ability to judge the part of speech (POS) of the word (e.g. ``ban.v.02'' gets larger weights than ``ban.n.04'').

\begin{wraptable}{r}{8.3cm}
\setlength{\belowcaptionskip}{-6mm}
\centering
\scriptsize
 \begin{tabular}{ccc}
  \toprule
  \makecell[c]{\textbf{Word in text}\\\textbf{(prototype)}} & \makecell[c]{\textbf{The most relevant}\\ \textbf{mentioned entity in}\\ \textbf{WordNet (predicted)}} & \textbf{Golden mentioned entity}\\ \hline
  ford &  ford.n.05 (0.56)	&  ford.n.05	\\ \hline
  pardon &  pardon.v.02 (0.86)	& pardon.v.02	\\ \hline
  nixon &  nixon.n.01 (0.74) & nixon.n.01	\\ \hline
  lead &  lead.v.03 (0.73) & lead.v.03		\\ \hline
  outrage &  outrage.n.02 (0.62) & outrage.n.02 	\\ 
\bottomrule
\end{tabular}
\caption{Case study. Comparisons between the golden label with the most relevant mentioned entity in WordNet. The importance of selected mentioned entities is provided in the parenthesis.}
\label{case study v2}
\end{wraptable}

For KELM, by contrast, we focus on selecting the most relevant mentioned entities to solve the knowledge ambiguity issue (based on the ``entity link to token graph'' part of UKET). For injecting WordNet, by allowing message passing on the extracted sub-graphs (``KG only'' part of UKET), knowledge context can be dynamically embedded according to the textual context. Thus the neighbors' information of mentioned entities in WordNet can be used to help the word in a text to correspond to a particular POS based on its context. The top 3 most relevant mentioned entities in WordNet for the word ``ban'' in the above example are ``ban.n.04'', ``ban.v.02'', and ``ban.v.01'', with the weights of 0.715, 0.205, and 0.060, respectively.

To vividly show the effectiveness of KELM, we analyze ambiguous words in the motivating example show in Fig.~\ref{KWG} (The example comes from ReCoRD): 

``\textit{President Ford then pardoned Richard Nixon, leading to a further firestorm of outrage.}'' 

Table.~\ref{case study v2} presents 5 words in the above passage. For each word, the most relevant mentioned entity in WordNet with the highest score is given. The golden mentioned entity for each word is labeled by us.
Definitions of mentioned entities in WordNet that correspond to the word examples are listing in Table~\ref{case study def} of Appendix.

\section{Conclusion}

In this paper, we have proposed KELM for MRC, which enhances PLM representations with structured knowledge from KGs based on the fine-tuning process. Via a unified knowledge-enhanced text graph, KELM can embed the injected knowledge dynamically, and select relevant mentioned entities in the input KGs. In the empirical analysis, KELM shows the effectiveness of fusing external knowledge into representations of PLM and demonstrates the ability to avoid knowledge ambiguity issue. 
Injecting emerging factual knowledge into PLM during finetuning without re-pretraining the whole model is quite important in the application of PLMs and is still barely investigated. Improvements achieved by KELM over vanilla baselines indicate a potential direction for future research.

\section*{Acknowledgements}
The authors thank Ms. X. Lin for insightful comments on the manuscript. We also thank Dr. Y. Guo for helpful suggestions in parallel training settings. We also thank all the colleagues in AI Application Research Center (AARC) of Huawei Technologies for their supports.
\bibliography{iclr2022_conference}
\bibliographystyle{iclr2022_conference}

\appendix
\section{Summary and Comparison of Recent Knowledge-enhanced PLMs}
\label{sec_pre:appendix}

Table~\ref{model comparison v1} shows a brief summary and comparison of recent knowledge-enhanced PLMs. Most of recent work concentrated on injecting external knowledge graphs during pre-training phase, which makes them inefficient in injecting external knowledge (e.g. LUKE takes about 1000 V100 GPU days to re-pretraining the RoBERTa based PLM model). Also, nearly all of them uses an additional entity linking tool to align the mentioned entities in the Wikidata to the entity mentions in the pre-trained corpus (English Wikipedia) uniquely. These methods never consider to resolve the knowledge ambiguity problem.

\begin{table*}[!ht]
\setlength{\belowcaptionskip}{-4mm}
\centering
\resizebox{\textwidth}{38mm}{
\begin{tabular}{lccccccccc}
     \toprule
  \textbf{Model} & \textbf{Downstream Task} & \textbf{Used KGs} & \makecell[c]{ \textbf{Need} \\ \textbf{Pre-train}}  &  \makecell[c]{ \textbf{Dynamically} \\ \textbf{Embedding KG} \\ \textbf{Context}}   & \makecell[c]{\textbf{Inject external} \\ \textbf{KG's Representations}} & \makecell[c]{\textbf{Support} \\ \textbf{Multi-relational}}  & \makecell[c]{ \textbf{Support} \\ \textbf{Multi-hop}}  & \makecell[c]{ \textbf{Handle Knowledge} \\ \textbf{Ambiguity Issue}} & \makecell[c]{ \textbf{Base Model}}\\ \hline 
    
 ERNIE~\citep{zhang2019ernie}  & \makecell[c]{Glue, Ent Typing \\ Rel CLS}  & Wikidata  & \makecell[c]{Yes\\(MLM, NSP, \\Ent Mask task)}  &  No & \makecell[c]{Inject pretrained\\ entity embeddings\\ (TransE) explicitly}  & \makecell[c]{No \\(only entity embedding)}  & No  & \makecell[c]{No \\(anchored entity mention to\\the unique id of Wikidata)} & \(\text{BERT}_{\text{base}}\)   \\ \hline
 
 K-BERT~\citep{liu2019kbert} & \makecell[c]{Q\&A, NER \\ Sent CLS}  & \makecell[c]{CN-DBpedia \\ HowNet, MedicalKG}  & \makecell[c]{\textbf{Optional}\\(MLM, NSP)}  & No  & No  & \textbf{\makecell[c]{Yes \\(treat relations as words)}}  & No  & \makecell[c]{No \\(designed ATT mechanism\\ can solve KN issue)} & \(\text{BERT}_{\text{base}}\)  \\ \hline
 
 KnowBERT~\citep{peters2019knowledge}  & \makecell[c]{Rel Extraction \\ Ent Typing}   & \makecell[c]{CrossWikis, \\WordNet}  & \makecell[c]{Yes\\(MLM, NSP, \\Ent Linking task)}   & No  & \makecell[c]{Inject both pretrained\\ entity embeddings (TuckER)\\ and entity definition explicitly}  & \makecell[c]{No \\(only entity embedding)}  & No  & \textbf{\makecell[c]{Yes \\(weighed entity embeddings\\shared the same text)}} & \(\text{BERT}_{\text{base}}\)  \\ \hline
 
 WKLM~\citep{xiong2019pretrained}   & \makecell[c]{Q\&A, Ent Typing}  & Wikidata  & \makecell[c]{Yes\\(MLM, Ent \\replacement task)}  & No  & No  & No  & No  & \makecell[c]{No \\(anchored entity mention to\\the unique id of Wikidata)}  & \(\text{BERT}_{\text{base}}\) \\  \hline
 
 
 K-Adapter~\citep{wang2020kadapter} & \makecell[c]{Q\&A, \\Ent Typing}  & \makecell[c]{Wikidata \\ Dependency Parsing}  & \makecell[c]{Yes\\(MLM, \\Rel predition task)}  & No  & No  & \textbf{\makecell[c]{Yes\\(Via Rel prediction task \\during pretraining)}}  & No  & \makecell[c]{No \\(anchored entity mention to\\the unique id of Wikidata)}  & \(\text{RoBERTa}_{\text{large}}\) \\ \hline
 
 KEPLER~\citep{wang2020kepler}  & \makecell[c]{Ent Typing \\ Glue, Rel CLS \\ Link Prediction}   & Wikidata  & \makecell[c]{Yes\\(MLM, \\Link predition task)}  & \textbf{Yes}  & \makecell[c]{Inject embeddings of \\entity and relation's \\description explicitly}  & \textbf{\makecell[c]{Yes\\(Via link prediction task\\during pretraining)}}  & No  & \makecell[c]{No \\(anchored entity mention to\\the unique id of Wikidata)} & \(\text{RoBERTa}_{\text{base}}\)\\ \hline
 
 
 JAKET~\citep{yu2020jaket}  & \makecell[c]{Rel CLS, KGQA \\ Ent CLS}  & Wikidata  & \makecell[c]{Yes\\(MLM, Ent Mask task, \\ Ent category prediction, \\ Rel type prediction)}  &  \textbf{Yes} & \makecell[c]{Inject embeddings of \\ entity descriptions}  & \textbf{\makecell[c]{Yes\\(Via Rel type prediction \\during pretraining)}}  & \textbf{Yes}  & \makecell[c]{No \\(anchored entity mention to\\the unique id of Wikidata)} & \(\text{RoBERTa}_{\text{base}}\)  \\ \hline
 
 CoLAKE~\citep{sun2020colake} & \makecell[c]{Glue, Ent Typing \\ Rel Extraction}  & Wikidata  & \makecell[c]{Yes\\(MLM, Ent Mask task, \\ Rel type prediction)}  & \textbf{Yes}  & No  & \textbf{\makecell[c]{Yes \\(treat relations as words)}}  & No  & \makecell[c]{No \\(anchored entity mention to\\the unique id of Wikidata)}  & \(\text{RoBERTa}_{\text{base}}\)\\ \hline
 
 LUKE~\citep{yamada2020luke} & \makecell[c]{Ent Typing, Rel CLS\\NER, Q\&A}  & \makecell[c]{Ent from\\ Wikipedia}  & \makecell[c]{Yes\\(MLM, Ent Mask task)}  & No  & No  & No & No  & \makecell[c]{No \\(treat hyperlinks in Wikipedia\\ as entity annotations)}  & \(\text{RoBERTa}_{\text{large}}\)\\ \hline
 
 CokeBERT~\citep{su2020cokebert}  & \makecell[c]{Rel CLS \\ Ent Typing}  & Wikidata  & \makecell[c]{Yes\\(MLM, NSP, \\Ent Mask task)}  & \textbf{Yes}  & \makecell[c]{Inject pretrained\\ entity embeddings\\ (TransE) explicitly}  & \textbf{\makecell[c]{Yes\\(Via S-GNN to encode KG \\context dynamically)}} & \textbf{Yes}  & \makecell[c]{No \\(anchored entity mention to\\the unique id of Wikidata)}  & \(\text{RoBERTa}_{\text{large}}\)\\ \hline
 
 SKG~\citep{qiu-etal-2019-machine}   & MRC  & \makecell[c]{WordNet, ConceptNet}  & \textbf{No} & \textbf{Yes}  & \makecell[c]{Inject pretrained\\ entity embeddings\\ (BILINER) explicitly}  & \textbf{\makecell[c]{Yes\\(Via multi-relational\\ GNN to encode KG\\ context dynamically)}}  & \textbf{Yes}  & No & \(\text{BERT}_{\text{large}}\) \\ \hline
 
 KT-NET~\citep{yang-etal-2019-enhancing-pre}  & MRC   & WordNet, NELL  & \textbf{No} & No  & \makecell[c]{Inject pretrained\\ entity embeddings\\ (BILINER) explicitly}  & \makecell[c]{No \\(only entity embedding)}  & No  & \textbf{\makecell[c]{Yes \\(dynamically selecting\\ KG context)}} & \(\text{BERT}_{\text{large}}\) \\ \hline
 
 \textbf{KELM}   & MRC  & WordNet, NELL  & \textbf{No}  & \textbf{Yes}  & \makecell[c]{Inject pretrained\\ entity embeddings\\ (BILINER) explicitly}  & \textbf{\makecell[c]{Yes\\(Via multi-relational \\GNN to encode KG\\ context dynamically)}}  & \textbf{Yes}  & \textbf{\makecell[c]{Yes \\(dynamically selecting\\ related mentioned entity)}} & \(\text{BERT}_{\text{large}}\)\\ \hline
  \end{tabular} 
}
\caption{A brief summary and comparison of recent knowledge-enhanced PLMs. The full names of some abbreviations are as follows. \textbf{MLM}: masked language model, \textbf{NSP}: next sentence prediction, \textbf{Ent}: entity, \textbf{Rel}: relation, \textbf{CLS}: classification, \textbf{Sent}: sentence, \textbf{ATT}: attention. Comments/descriptions of features are written in parentheses. Desired properties are written in \textbf{bold}.}
\label{model comparison v1}
\end{table*}

\section{Datatset and Knowledge Graph Details}
\label{secA:appendix}
\textbf{ReCoRD} (an acronym for the Reading Comprehension with Commonsense Reasoning Dataset) is a large-scale dataset for extractive-style MRC requiring commonsense reasoning. There are 100,730, 10,000, and 10,000 examples in the training, development (dev), and test set, respectively. An example of the ReCoRD consists of three parts: passage, question, and answer. The passage is formed by the first few paragraphs of an article from CNN or Daily Mail, with named entities recognized and marked. The question is a sentence from the rest of the article, with a missing entity specified as the golden answer. The model needs to find the golden answer among the entities marked in the passage. 
Questions that can be easily answered by pattern matching are filtered out.
By the design of the process of data collection, one can see that to answer the questions, external background knowledge and ability of reasoning are required. 

\noindent \textbf{MultiRC} (Multi-Sentence Reading Comprehension) is a multiple-response-items-style MRC dataset of short paragraphs and multi-sentence questions that can be answered from the content of the paragraph. 
Each example of MultiRC includes a question that associates with several choices for answer-options, and the number of correct answer-options is not pre-specified. The correct answer is not required to be a span in the text. 
The dataset consists of ~10K questions (~6k multiple-sentence questions), about 60\% of this data make training/dev data.
Paragraphs in the dataset have diverse provenance by being extracted from 7 different domains such as news, fiction, historical text etc., and hence are expected to be more complicated in their contents as compared to single-domain datasets.

\noindent \textbf{WordNet} contains 151,442 triplets with 40,943 synsets459and 18 relations. We look up mentioned entities in the WordNet by string matching operation, and link all tokens in the same word to the retrieved mentioned entities (tokens are tokenized by Tokenizer of BERT). Then, we extract all 1-hop neighbors for each mentioned entity and construct sub-graphs. In this paper, our experiment results are based on the 1-hop case. However, our framework can be generalized to multi-hop easily, and we leave this for future work.

\noindent \textbf{NELL} contains 180,107 entities and 258 concepts. We link entity mentions to the whole KG, and return associated concepts.

\section{Implementation Details}
\label{secB:appendix}
Our implementation is based on HuggingFace~\citep{wolf2020huggingfaces} and DGL~\citep{wang2020deep}. For all three settings of KELM, parameters of the encoding layer of \(\text{BERT}_{\text{large}}\) are initialized with pre-trained model released by Google. Other trainable parameters in HMP are randomly initialized. The total number of trainable parameters of KELM is 340.4M (Roughly the same as \(\text{BERT}_{\text{large}}\), which has 340M parameters). 
Since including all neighbors around mentioned entities of WordNet is not efficient, for simplicity, we use top 3 most common relations in WordNet in our experiment (i.e. hyponym, hypernym, derivationally\_related\_form). For both datasets, we use a ``two stage'' fine-tune strategy to achieve our best performance, the FullTokenizer built in BERT is used to segment input words into wordpieces. 

For ReCoRD, the maximum length of answer during inference is set to 30, and the maximum length of question is set to 64. Questions longer than that are truncated. The maximum length of input sequence \(T\)$\footnote{Refer to the PLM Encoding Module of Methodology Section.}$ is set to 384. Input sequences longer than that are segmented into chunks with a stride of 128. Fine-tuning our model on ReCoRD costs about 18 hours on 4 V100 GPU with a batch size of 48. We freeze parameters of BERT and use Adam optimizer with a learning rate of 1e-3 to train our knowledge module in the first stage. The maximum number of training epochs of the first stage is 10. The purpose of this is to provide a good weight initialization for our HMP. For the second stage, the pre-trained BERT parameters and our HMP part will be fine-tuned together. The max number of training epochs is chosen from \(\left\{4, 6, 8\right\}\). The learning rate is set to be 2e-5 with a warmup over the first \(6\%\) of max steps, and linear decay until up to max epochs. For both two stages, early stopping is applied according to the best EM+F1 score on the dev set every 500 steps. 

For MultiRC, the maximum length of input sequence \(T\) is set to 256. The summation of length of question (\(Q\)) and length of candidate answer (\(A\)) is not limited. Paragraph (\(P\)) is truncated to fit the maximum length of input sequence. Fine-tuning KELM on MultiRC needs about 12 hours on 4 V100 GPU with a batch size of 48. For the first stage finetuning, learning rate is 1e-4 and the maximum number of training epochs is 10. For the second stage, the max number of training steps is chosen from \(\left\{10000, 15000, 20000\right\}\). The learning rate is set to be 2e-5 with a warmup over the first \(10\%\) of max steps.

\section{Supplementation of the Case Study Section}
\label{secC:appendix}
We provide definitions of the top 3 most relevant mentioned entities in WordNet that correspond to the word examples mentioned in \textbf{Case Study Section}. Descriptions are obtained by using NLTK~\citep{10.3115/1118108.1118117}. By comprehending the motivating example in the case study section, we can see that KELM can correctly select the most relevant mentioned entities in the KG. 

\begin{table}[!ht]
\centering
\tiny
\setlength{\tabcolsep}{0.5mm}{
 \begin{tabular}{ccl}
  \toprule
    \makecell[c]{\textbf{Word in text}\\\textbf{(prototype)}} & \makecell[c]{\textbf{Mentioned entity}\\ \textbf{in WordNet}} & \textbf{Definition}  \\ 
  \midrule
    \multirow{3}*{ban}	
    ~   &	ban.n.04 (0.72)	&	an official prohibition or edict against something	\\
    ~	&	ban.v.02 (0.21)	&	prohibit especially by legal means or social pressure	\\
    ~	&	ban.v.01 (0.06)	& forbid the public distribution of ( a movie or a newspaper)	\\
    \midrule															
    \multirow{5}*{ford} 
    ~   &	ford.n.05 (0.56)	&	\makecell[l]{38th President of the United States; \\appointed vice president and succeeded \\Nixon when Nixon resigned (1913-)}	\\
    ~	&	ford.n.07 (0.24)	&	a shallow area in a stream that can be forded	\\
    ~	&	ford.v.01 (0.08)	& cross a river where it's shallow	\\
    \midrule	
    \multirow{3}*{pardon}	
    ~   &	pardon.v.02 (0.86)	&	a warrant granting release from punishment for an offense		\\
    ~	&	sentinel (0.10)	&	-	\\
    ~	&	pardon.n.02 (0.04)	& grant a pardon to	\\
    \midrule	
    \multirow{3}*{nixon}	
    ~   &	nixon.n.01 (0.74)	&	\makecell[l]{vice president under Eisenhower and 37th President\\ of the United States; resigned after the Watergate \\scandal in 1974 (1913-1994)}	\\
    ~	&	sentinel (0.26)	&	-	\\
    \midrule	
    \multirow{3}*{lead}	
    ~   &	lead.v.03 (0.73)	&	tend to or result in		\\
    ~	&	lead.n.03 (0.12)	&	evidence pointing to a possible solution	\\
    ~	&	lead.v.04 (0.05)	&   travel in front of; go in advance of others	\\
    \midrule	
    \multirow{2}*{outrage}
    ~   &	outrage.n.02 (0.62)	&	a wantonly cruel act	\\
    ~	&	sentinel (0.38)	&	- \\

\bottomrule
\end{tabular} }
\caption{Definitions of mentioned entities in WordNet corresponding to the word examples in the case study. The importance of mentioned entities is provided in the parenthesis. ``sentinel'' is meaningless, which is used to avoid knowledge noise.}
\label{case study def}
\end{table}

\section{Experiment on Commonsense Causal Reasoning Task} 
\label{secD:appendix}
To further explore the generalization ability of KELM, we also evaluate our method on \textbf{COPA}~\citep{DBLP:conf/aaaiss/RoemmeleBG11} (Choice of Plausible Alternatives), which is also a benchmark dataset in SuperGLUE and can be used for evaluating progress in open-domain commonsense causal reasoning. COPA consists of 1000 questions, split equally into development and test sets of 500 questions each. Each question is composed of a premise and two alternatives, where the task is to select the alternative that more plausibly has a causal relation with the premise. Similar to the previous two MRC tasks, the development set is publicly available, but the test set is hidden. One has to submit the predicted results for the test set to SuperGLUE to retrieve the final test score.
Since the implementation of KELM is based on \(\text{BERT}_{\text{large}}\), we use it as our baseline for the comparison. The result of \(\text{BERT}_{\text{large}}\) is directly taken from the leaderboard of SuperGLUE. Table~\ref{copa table} shows the experiment results. The injected KG is WordNet here, and we use accuracy as the evaluation metric. 

\begin{table}[!ht]
\centering
\scriptsize
 \begin{tabular}{ccc}
  \toprule
     \textbf{Model} & \textbf{dev}	&	\textbf{test}  \\
  \midrule
    \(\text{BERT}_{\text{large}}\)	&	-	&	70.6	\\
    \midrule													
    \(\text{KELM}^{\text{BERT}_{\text{large}}}_{\text{WordNet}}\)	&	\textbf{76.1}	&	\textbf{78.0}	\\
\bottomrule
\end{tabular} 
\caption{Performance comparison on COPA. The effectiveness of injecting knowledge (WordNet) are shown.}
\label{copa table}
\end{table}

The huge improvement over the baseline in this task demonstrates that knowledge in WordNet is indeed helpful for BERT to improve the generalization ability to the out-of-domain downstream task.

\section{KELM: A framework of finetune-based model-agnostic knowledge-enhanced PLM} 
\label{secE:appendix}
We implement KELM based on the \(\text{RoBERTa}_{\text{large}}\), which has a similar number of trainable parameters as \(\text{BERT}_{\text{large}}\) but uses nearly 10 times of training corpus than \(\text{BERT}_{\text{large}}\). 
Since the performances of RoBEATa on the leaderboard of SuperGLUE are based on ensembling, we also finetune \(\text{RoBERTa}_{\text{large}}\) on ReCoRD to produce the results of a single model. 
Comparisons of the results can be found in Table~\ref{roberta tabel}, where you can also see an improvement there. However, that improvement is not as significant as we observed in \(\text{BERT}_{\text{large}}\). 
Reasons are two-fold:
(1) Passages in ReCoRD are collected from articles in CNN/Daily Mail, while BERT is pre-trained on BookCorpus and English Wikipedia. RoBERTa not only uses the corpus that used in BERT (16 GB), but also an additional corpus collected from the CommonCrawl News dataset (76 GB). 
ReCoRD dataset is \textbf{in-domain} for RoBERTa but is \textbf{out-of-domain} for BERT. \textbf{It seems that the improvements of KELM with injecting general KGs (e.g. WordNet) on the in-domain downstream tasks are not as large as the out-of-domain downstream tasks}. A similar phenomenon can be also observed in the experiment of SQuAD 1.1 (Refer to Appendix E).
(2) The same external knowledge (WordNet, NELL) can not help \(\text{RoBERTa}_{\text{large}}\) too much, since RoBERTa is pre-trained on a much larger corpus than BERT, knowledge in WordNet/NELL has been learned in RoBERTa.

\begin{table}[!ht]
\centering
\scriptsize
 \setlength{\tabcolsep}{1.2mm}
 \begin{tabular}{clcccc}
  \toprule
    \multicolumn{2}{c}{} & \multicolumn{2}{c}{\textbf{Dev}} & \multicolumn{2}{c}{\textbf{Test}}  \\
  	&	\textbf{Model} & \textbf{EM}	&	\textbf{F1} &	\textbf{EM}	&	\textbf{F1} \\
  \midrule
    \multirow{2}*{\makecell[c]{ \textbf{PLM w/o} \\\textbf{external knowledge}}}	&	\(\text{BERT}_{\text{large}}\)	&	70.2	&	72.2 & 71.3 & 72.0	\\
    ~	&	\(\text{RoBERTa}_{\text{large}}^{*}\)	&	87.9	&	88.4 & 88.4 & 88.9	\\
    \midrule													
    \multirow{2}*{\makecell[c]{\textbf{knowledge enhanced PLM}\\ \textbf{(finetune-based)}\\}}	
    ~	&	\(\text{KELM}^{\text{BERT}_{\text{large}}}_{\text{Both}}\)	&	\textbf{75.1}	&	\textbf{75.6} & \textbf{76.2} & \textbf{76.7}	\\
    ~	&	\(\text{KELM}^{\text{RoBERTa}_{\text{large}}}_{\text{Both}}\)	&	\underline{88.2}	&	\underline{88.7} & \underline{89.1} & \underline{89.6}	\\
    \midrule												
    \multirow{2}{*}{\makecell[c]{\textbf{knowledge enhanced PLM}\\ \textbf{(pretrain-based)}\\}}	
    ~	&	LUKE	&	\textbf{90.8}	&	\textbf{91.4} & \textbf{90.6} & \textbf{91.2}	\\
    ~   & & & & &\\
\bottomrule
\end{tabular} 
\caption{Comparison of the effectiveness of injecting external knowledge between BERT and RoBERTa. [*] Results are from our implementation.}
\label{roberta tabel}
\end{table}

We also list the results of LUKE~\citep{yamada2020luke} in Table~\ref{roberta tabel}. LUKE is a \textbf{pretrain-based} knowledge enhanced PLM and uses Wiki-related golden entities (one-to-one mapping) as the injected knowledge source (about 500k entities$\footnote{For KELM, we only use 40943 entities in WordNet and 258 concepts in NELL.}$). It has more 128 M parameters than the total number of parameters of the vanilla RoBERTa. As we summarized in Table 4 in the main text, the pre-training task is also different compared with RoBERTa. Although LUKE gets better results compared with vanilla RoBERTa and KELM, it needs 16 NVIDIA Tesla V100 GPUs and the training takes approximately 30 days. Relying on hyperlinks in Wikipedia as golden entity annotations, lacking the flexibility to adapt the external knowledge of other domains, and needing re-pretraining when incorporating knowledge, these limitations hinder the abilities of applications. 

\section{Experiment on SQuaD 1.1} 
\label{secF:appendix}
SQuAD1.1~\citep{rajpurkar2016squad} is a well known extractive-style MRC dataset that consists of questions created by crowdworkers for \textbf{Wikipedia articles}. It contains 100,000+ question-answer pairs on 536 articles. We implement KELM based on the \(\text{BERT}_{\text{large}}\), and compare our results on the development set of SQuAD 1.1 with KT-NET (Best result of KT-NET is based on injecting WordNet only). Results are shown in Table~\ref{squad table}

\begin{table}[!ht]
\centering
\scriptsize
 \setlength{\tabcolsep}{1.2mm}
 \begin{tabular}{clcc}
  \toprule
    \multicolumn{2}{c}{} & \multicolumn{2}{c}{\textbf{Dev}} \\
  	&	\textbf{Model} & \textbf{EM}	&	\textbf{F1}  \\
  \midrule
    \multirow{1}*{\makecell[c]{ \textbf{PLM w/o} \textbf{external knowledge}}}	&	\(\text{BERT}_{\text{large}}\)	&	84.4	&	91.2	\\
    \midrule													
    \multirow{2}*{\makecell[c]{\textbf{knowledge enhanced PLM}\\ \textbf{(finetune-based)}\\}}	
    ~	&	\(\text{KT-NET}^{\text{BERT}_{\text{large}}}_{\text{WordNet}}\)	&	\textbf{85.1}	&	\textbf{91.7} 	\\
    ~	&	\(\text{KELM}^{\text{BERT}_{\text{large}}}_{\text{WordNet}}\)	&	\underline{84.7}	&	\underline{91.5} \\
\bottomrule
\end{tabular} 
\caption{Performance comparison on the development set of SQuAD 1.1.}
\label{squad table}
\end{table}

Results on KELM show an improvement over vanilla BERT. Both BERT and RoBERTa use English Wikipedia as the corpus for pretraining. Since SQuAD is also created from Wikipedia, it is an in-domain downstream task for both BERT and RoBERTa (while ReCoRD dataset is \textbf{in-domain} for RoBERTa but is \textbf{out-of-domain} for BERT). This explains why RoBERTa achieves a much larger improvement over BERT on the result of ReCoRD (\(71.3\rightarrow 88.4\) in EM on test set) than the one on SQuAD 1.1 (\(84.1\rightarrow88.9\)). The rest of the improvement is because RoBERTa uses 10 times of training corpus than BERT and different pre-training strategies they used. 

Interestingly, we find the performance of KELM on SQuAD 1.1 is sub-optimal compared with KT-NET. 
As we mentioned in the last paragraph of the \textbf{Related Work Section}, KT-NET treats all synsets of entity mentions within the WN18 as candidate KB concepts. Via a specially designed attention mechanism, KT-NET can directly use all 1-hop neighbors of the mentioned entities. Although this limits the ability of KT-NET to select the most relevant mentioned entities (as we discussed in \textbf{Case Study Section}), information of these neighbors can be directly considered. 
Using neighbors of the mentioned entities indirectly via the HMP mechanism makes it possible for KELM to dynamically embed injected knowledge and to select semantics-related mentioned entities. 
However, SQuAD is an in-domain downstream task for BERT, the problem of ambiguous meanings of words can be alleviated by pretraining model on the in-domain corpus. Compared with KT-NET, a longer message passing path in KELM may lead to sub-optimal improvement on the in-domain task.



\section{Further Discussions About the Novelty w.r.t SKG/KT-NET} 
\label{secG:appendix}
UKET defined in KELM consists of three subgraphs in a hierarchical structure, each subgraph corresponds to one sub-process of our proposed HMP mechanism and solves one problem presented in the \textbf{Hierarchical Knowledge Enhancement Module} part of \textbf{Methodology Section}. SKG only uses GNN to dynamically encode the extracted KG which corresponds to the first part of UKET, it can not solve the knowledge ambiguity issue and forbids interactions among knowledge-enriched tokens. KT-NET defines a similar graph as the third part of UKET. However, the first and second subgraphs of UKET are absent. The second subgraph of UKET is independent of ideas of KT-NET and SKG, thus KELM is not a simple combination of these two methods. We are the \textbf{first to unify text and KG into a graph} and to propose this hierarchical message passing framework to incorporate two heterogeneous information. \textbf{SKG/KT-NET can be interpreted as parts of the ablation study of components of KELM}. The result of SKG is ablation with the component only related to the first subgraph of UKET. While KT-NET only contains the third subgraph with a modified knowledge integration module. KELM uses a dedicatedly designed HMP mechanism to let the information of farther neighbors to be considered. However, longer information passing path than KT-NET makes it less efficient. In our experiments, KELM takes more 30\% training time than KT-NET on both ReCoRD and MultiRC.

\section{Limitations and Further Improvements of KELM} 
\label{secH:appendix}
Limitations for KELM are two-fold: (1) Meanings of mentioned entities in different KGs that share the same entity mentions in the text may conflict with each other. Although HMP can help to select the most relevant mentioned entities in a single KG, there's no mechanism to guarantee the selections across different KGs; (2) Note the knowledge-enriched representation in Eq.3 is obtained by simple concatenation of the embeddings from different KGs. Too much knowledge incorporation may divert the sentence from its correct meaning (Knowledge noise issue). We expect these two potential improvements to be a promising avenue for future research.

\section{Further Analysis and Discussion}
\label{secI:appendix}
KELM incorporates knowledge in KGs into the representations in the last hidden layer of PLM (Refer to \textbf{Methodology Section}). It is essentially a model-agnostic, KG-agnostic, and task-agnostic framework for enhancing language model representations with factual knowledge from KGs. It can be used to enhance any PLM, with any injected KGs, on any downstream task. 
Besides the two Q\&A-related MRC tasks we mentioned in the main text, we also evaluate KELM on COPA and SQuAD 1.1 based on \(\text{BERT}_{\text{large}}\), results are presented in Appendix E and Appendix G, respectively. To demonstrate KELM is a model-agnostic framework, we also implement KELM based on \(\text{RoBERTa}_{\text{large}}\) and evaluate it on ReCoRD. The experiment is presented in Appendix F. Improvements achieved by KELM over all vanilla base PLM models indicate the effectiveness of injecting external knowledge.

However, the improvements of KELM over RoBERTa on ReCoRD and BERT on SQuAD 1.1 are marginal compared with the ones on ReCoRD/MultiRC/COPA (\(\text{BERT}_{\text{large}}\) based). The reason behind this is that pretraining model on \textbf{in-domain} unlabeled data could boost performance on downstream tasks. 
Passages in ReCoRD are collected from articles in CNN/Daily Mail, while BERT is pre-trained on BookCorpus and English Wikipedia. RoBERTa not only uses the corpus that used in BERT (16 GB), but also an additional corpus collected from the CommonCrawl News dataset (76 GB). ReCoRD is \textbf{in-domain} for RoBERTa but is \textbf{out-of-domain} for BERT. 
Similarly, SQuAD 1.1 is created from Wikipedia, it is an \textbf{in-domain} downstream task for both BERT and RoBERTa. This partially explains why RoBERTa achieves a much larger improvement over BERT on the result of ReCoRD (\(71.3\rightarrow 88.4\) in EM on test set) than the one on SQuAD 1.1 (\(84.1\rightarrow88.9\)). A similar analysis can be also found in T5~\citep{raffel2020exploring}.
From our empirical results, we can summarize that general KG (e.g. WordNet) can not help too much for the PLMs pretrained on in-domain data. But it can still improve the performance of the model when the downstream tasks are out-of-domain. Further detailed analysis can be found in our appendix.

Finding a popular NLP task/dataset that is not related to the training corpus of modern PLMs is difficult. Pre-training on large-scale corpus is always good if we have unlimited computational resources and plenty of in-domain corpus. It has been evident that the simple finetuning of PLM is not sufficient for domain-specific applications. KELM can provide people another choice when they do not have such a large-scale in-domain corpus and want to incorporate incremental domain-related structural knowledge into the domain-specific applications.

\end{document}